# TSONN: Time-stepping-oriented neural network for solving partial differential equations

Wenbo Cao, Weiwei Zhang*

*School of Aeronautics, Northwestern Polytechnical University, Xi'an 710072, China.*

Deep neural networks (DNNs), especially physics-informed neural networks (PINNs), have recently become a new popular method for solving forward and inverse problems governed by partial differential equations (PDEs). However, these methods still face challenges in achieving stable training and obtaining correct results in many problems, since minimizing PDE residuals with PDE-based soft constraint make the problem ill-conditioned. Different from all existing methods that directly minimize PDE residuals, this work integrates time-stepping method with deep learning, and transforms the original ill-conditioned optimization problem into a series of well-conditioned sub-problems over given pseudo time intervals. The convergence of model training is significantly improved by following the trajectory of the pseudo time-stepping process, yielding a robust optimization-based PDE solver. Our results show that the proposed method achieves stable training and correct results in many problems that standard PINNs fail to solve, requiring only a simple modification on the loss function. In addition, we demonstrate several novel properties and advantages of time-stepping methods within the framework of neural network-based optimization approach, in comparison to traditional grid-based numerical method. Specifically, explicit scheme allows significantly larger time step, while implicit scheme can be implemented as straightforwardly as explicit scheme.

## 1 Introduction

As a scientific machine learning technique, deep neural network (DNN) represented by physics-informed neural networks (PINNs) [1] has recently been widely used to solve forward and inverse problems involving partial differential equations (PDEs). By minimizing the loss of PDE residuals, boundary conditions and initial conditions simultaneously, the solution can be straightforwardly obtained without mesh, spatial discretization, and complicated program. The concept of PINNs can be traced back to 1990s, where neural algorithms for solving differential equations were proposed [2-6]. With the significant progress in deep learning and computation capability, a variety of PINN models have been proposed in the past few years, and have achieved remarkable results across a range of problems in computational science and engineering [7-10]. As a typical optimization-based PDE solver, PINNs offer a natural approach to solve PDE-constrained optimization problems. A promising application is on the flow visualization technology [11-13], where the flow fields can be easily inferred from sparse observations such as concentration fields and images. Such inverse problems are difficult for traditional PDE solvers. Moreover, several recent works [14-17] have demonstrated the significant advantages of PINNs in inverse design and optimal control, where solving the PDE and obtaining optimal design are pursued simultaneously. In contrast, the most popular method, direct-adjoint-looping (DAL), involves hundreds or thousands of repeated evaluations of the forward PDE, yielding prohibitively expensive computational cost in solving PDE-constrained optimization problems.

Despite the potential for a wide range of physical phenomena and applications, training PINN models still encounters challenges [7], and the current generation of PINNs is not as accurate or as efficient as traditional numerical method for solving many forward problems [18, 19]. There have been some efforts to improve PINN trainability, which include adaptively balancing the weights of loss components during training [20-23], enforcing boundary conditions [6, 17, 24], transforming the PDE to its weak form [25-27], adaptively sampling [28, 29], spatial-temporal decomposition [30-32], using both automatic differentiation and numerical differentiation to calculate derivatives [33], and so on. Nevertheless, PINNs are still difficult to achieve stable training and obtain correct results in many problems, since minimizing PDE residuals with PDE-based soft constraint make the problem ill-conditioned [34]. Considering this, different from all existing methods that directly minimize PDE residuals, we integrate time-stepping method with deep learning, and transforms the original optimization problem into a series of more well-conditioned problems over given pseudo time intervals.

In computational physics, time-stepping method is a classical and widely used method to solve both time-independent and time-dependent problems governed by PDEs. For time-independent problems $\mathcal{N}[u(x)] = 0$, a pseudo time derivative is introduced, leading to a time-dependent problem

$$\frac{\partial u}{\partial \tau} = \mathcal{N}[u(x)] \tag{1}$$

To obtain the steady-state solution, we start from an arbitrary initial condition and march the solution to a sufficiently large pseudo time under the given boundary conditions (Equation (2)). In this case, the final steady-state is the desired result, and time-stepping is simply a means to reach this state. Although it seems inefficient to introduce time as another independent variable, sometimes this is the only way to have a well-posed problem [35, 36] and hence is the most robust way to obtain the steady-state solution, especially in computational fluid mechanics (CFD).

$$u_{n+1}(x) = u_n(x) + \Delta \tau \cdot \mathcal{N}[u_n(x)]$$
$$u_{n+1}(x) = u_n(x) + \Delta \tau \cdot \mathcal{N}[u_{n+1}(x)]$$
(2)

For time-dependent problems $\mathcal{N}[u(x,t)] = 0$, pseudo time derivative can also be introduced in each physical time step and then one can compute the solution in steps of pseudo time until convergence, which is the classic dual-time method in CFD [37].

In this paper, we use neural network (NN) to continuously approximate the label $u_{n+1}(x)$ from explicit or implicit pseudo time-stepping, so that the training follows the pseudo time-stepping trajectory until it converges to the desired solution. The remainder of the paper is organized as follows. In Section 2, the basic framework of PINNs is introduced and then the TSONN method is proposed. Following that we report numerical results on various problems that standard PINNs failed to solve in Section 3. Finally, the paper is concluded in Section 4.

## 2 Methodology

2.1 Physics-informed neural networks

In this section, we briefly introduce the PINNs methodology. A typical PINN uses a fully connected DNN architecture to represent the solution $u$ of the dynamical system. The network takes the spatial $x \in \Omega$ and temporal $t \in [0,T]$ as the input and outputs the approximate solution $\hat{u}(x,t;\theta)$. The spatial domain usually has 1-, 2- or 3-dimensions in most physical problems, and the temporal domain may not exist for steady (time-independent) problems. The accuracy of the PINN outputs is determined by the network parameters $\theta$, which are optimized with respect to the PINN loss function during the training. To derive the PINN loss function, we consider $u$ to be mathematically described by differential equations of the general form:

$$\mathcal{N}[u(x,t)] = 0, x \in \Omega, t \in (0,T]$$
$$\mathcal{I}[u(x,0)] = 0, x \in \Omega$$
$$\mathcal{B}[u(x,t)] = 0, x \in \partial\Omega, t \in (0,T]$$
(3)

where $\mathcal{N}[\cdot]$, $\mathcal{I}[\cdot]$ and $\mathcal{B}[\cdot]$ are the PDE operator, the initial condition operator, and the boundary condition operator, respectively. Then the PINN training loss function is defined as

$$\begin{aligned}
\mathcal{L} &= \mathcal{L}_{PDE} + \lambda_{BC}\mathcal{L}_{BC} + \lambda_{IC}\mathcal{L}_{IC} \\
\mathcal{L}_{PDE} &= \left\| \mathcal{N}[\hat{u}(\cdot;\boldsymbol{\theta})] \right\|^2_{\Omega\times(0,T]} \\
\mathcal{L}_{IC} &= \left\| \mathcal{I}[\hat{u}(\cdot,0;\boldsymbol{\theta})] \right\|^2_{\Omega} \\
\mathcal{L}_{BC} &= \left\| \mathcal{B}[\hat{u}(\cdot;\boldsymbol{\theta})] \right\|^2_{\partial\Omega\times(0,T]}
\end{aligned} \quad (4)$$

The relative weights, $\lambda_{BC}$ and $\lambda_{IC}$ in Equation (4), control the trade-off between different components in the loss function. The PDE loss is computed over a finite set of $m$ collocation points $D=\{x_i,t_i\}_{i=1}^{m}$ during training, along with boundary condition loss and initial condition loss. The gradients in the loss are computed via automatic differentiation [38].

2.2 Time-stepping method

Time-stepping method can be generally classified into explicit and implicit schemes. In an explicit scheme, each difference equation contains only one unknown and therefore can be solved in a straightforward manner. An implicit scheme is one where the unknowns must be obtained by means of a simultaneous solution of the difference equations applied at all the grid points at a given time level (Equation (2)). Thus, implicit scheme usually involves local linearization of nonlinear operators $\mathcal{N}$ and simultaneously solves a large system of algebraic equations, resulting in a wide variety of CFD techniques [39]. In contrast, explicit scheme is simple to implement, but its time step $\Delta\tau$ is largely limited by stability constraints, thus resulting in long computer running times over a given time interval. Compared to explicit scheme, the implicit scheme is more complicated and requires larger computational cost and memory requirement at per time step, while remains stable with a much larger $\Delta\tau$, and can be more efficient in CFD. For more details on time-stepping method, please refer to [35].

In this study, we introduce pseudo time derivatives for both time-independent and time-dependent problems. Therefore, for the $n$th optimization step of the neural network $\hat{u}(\cdot;\boldsymbol{\theta}_n)$, the explicit and implicit time-stepping schemes are represented in Equation (5). To avoid confusion, we use $\boldsymbol{\theta}$ to represent the network parameters being optimized, and $\boldsymbol{\vartheta}_n$ to represent the replica of $\boldsymbol{\theta}$ at the $n$th step.

$$\bar{u}_{n+1} = \hat{u}(\cdot;\vartheta_n) + \Delta\tau \cdot \mathcal{N}[\hat{u}(\cdot;\vartheta_n)]$$
$$\bar{u}_{n+1} = \hat{u}(\cdot;\vartheta_n) + \Delta\tau \cdot \mathcal{N}[\bar{u}_{n+1}]$$
(5)

### 2.3 Time-stepping-oriented neural network

In TSONN, the network takes the same input and output as PINNs, but the loss of PDE residual $\mathcal{L}_{PDE}$ is replaced by $\mathcal{L}_{eTS}$ or $\mathcal{L}_{iTS}$, which is the approximation loss to the label $\bar{u}_{n+1}$ based on either explicit or implicit scheme. Boundary conditions and initial conditions remain soft constraints like PINNs.

At the $n$th optimization step, the approximation loss $\mathcal{L}_{eTS}$ of explicit time-stepping-oriented neural network (eTSONN) is defined as

$$\mathcal{L}_{eTS} = 1/\Delta\tau^2 \left\| \hat{u}(\cdot;\boldsymbol{\theta}) - \bar{u}_{n+1} \right\|^2_{\Omega\times(0,T]}$$
$$= 1/\Delta\tau^2 \left\| \hat{u}(\cdot;\boldsymbol{\theta}) - \hat{u}(\cdot;\vartheta_n) - \Delta\tau \cdot \mathcal{N}[\hat{u}(\cdot;\vartheta_n)] \right\|^2_{\Omega\times(0,T]}$$
(6)

The $1/\Delta\tau^2$ in $\mathcal{L}_{eTS}$ is to avoid the significant impact of the pseudo time step $\Delta\tau$ on the relative weights of different components in the loss function. To make the model training follow the trajectory of time-stepping, it needs $K$ steps to adequately approximate $\bar{u}_{n+1}$. After $K$ steps to optimize $\boldsymbol{\theta}$, we perform a step of explicit time-stepping, i.e.,

$$\vartheta_{n+1} = \boldsymbol{\theta}$$
$$\bar{u}_{n+2} = \hat{u}(\cdot;\vartheta_{n+1}) + \Delta\tau \cdot \mathcal{N}[\hat{u}(\cdot;\vartheta_{n+1})]$$
$$n \leftarrow n+1$$
(7)

then we enter the next one. The optimization of $\boldsymbol{\theta}$ by minimizing $\mathcal{L}_{eTS}$ is called inner iteration, while the stepping of $n$ is called outer iteration. Equation (6) shows that the NN is only used to approximate the label at each outer iteration. Such tasks are the most basic applications of neural network, which significantly reduce the difficulty of optimization compared with PINNs.

In an implicit time-stepping-oriented neural network (iTSONN), we use the NN $\hat{u}(\cdot;\boldsymbol{\theta})$ to approximate $\bar{u}_{n+1}$, which satisfy the implicit equation

$$\bar{u}_{n+1} = \hat{u}(\cdot;\vartheta_n) + \Delta\tau \cdot \mathcal{N}[\bar{u}_{n+1}]$$
(8)

Thus, the NN can simultaneously solve and implicitly approximate $\bar{u}_{n+1}$ by minimizing

$$\mathcal{L}_{iTS} = 1/\Delta\tau^2 \left\| \hat{u}(\cdot;\boldsymbol{\theta}) - \bar{u}_{n+1} \right\|^2_{\Omega\times(0,T]}$$
$$= 1/\Delta\tau^2 \left\| \hat{u}(\cdot;\boldsymbol{\theta}) - \hat{u}(\cdot;\vartheta_n) - \Delta\tau \cdot \mathcal{N}[\hat{u}(\cdot;\boldsymbol{\theta})] \right\|^2_{\Omega\times(0,T]}$$
(9)

The iTSONN also requires inner iterations to adequately minimize $\mathcal{L}_{iTS}$ in each outer iteration. After $K$ steps inner iteration to optimize $\theta$, $\hat{u}(\cdot;\vartheta_{n+1})$ is calculated with the latest $\theta$, then we enter the next outer iteration.

Equation (6) and Equation (9) are encouraging because of their unified form for both explicit and implicit schemes. In the traditional numerical methods, an implicit scheme demands a simultaneous solution of a large system of nonlinear equation, which is an exceptionally difficult task. Therefore, in practice, the nonlinear equation must be locally linearized to form a large system of linear algebraic equations, which is solved with larger computational cost and memory requirement at per time step. In contrast, in NN-based optimization approach, as shown in Equation (9), the implementation of the implicit scheme is natural and as straightforwardly as the explicit scheme, and there is almost no additional computational cost and memory requirement. In addition, we observe that PINN is a special case of iTSONN when the pseudo time step $\Delta\tau$ is large enough. Thus, compared to PINNs, iTSONN divides the original problem into a series of well-conditioned sub-problems over given pseudo time step. A similar idea is also employed in Newton iteration, where the pseudo time step is introduced to generate a well-conditioned problem at each Newton iteration step [40].

**Algorithm 1:** Unified framework of eTSONN, iTSONN and PINNs

**Input:** Initial $\theta$, $\vartheta_1 = \theta$, collocation points, outer iterations $N$, inner iterations $K$, pseudo time step $\Delta\tau$.

1: for $n = 1, 2, \cdots, N$ do
2:     for $k = 1, 2, \cdots, K$ do
3:         (a) Compute the total loss by

$$\mathcal{L}(\theta) = \lambda_{BC}\mathcal{L}_{BC} + \lambda_{IC}\mathcal{L}_{IC} + \begin{cases} \mathcal{L}_{PDE} = \|\mathcal{N}[\hat{u}(\cdot;\theta)]\|^2 & \text{if PINN;} \\ \mathcal{L}_{eTS} = 1/\Delta\tau^2 \|\hat{u}(\cdot;\theta) - \hat{u}(\cdot;\vartheta_n) - \Delta\tau \cdot \mathcal{N}[\hat{u}(\cdot;\vartheta_n)]\|^2 & \text{if eTSONN;} \\ \mathcal{L}_{iTS} = 1/\Delta\tau^2 \|\hat{u}(\cdot;\theta) - \hat{u}(\cdot;\vartheta_n) - \Delta\tau \cdot \mathcal{N}[\hat{u}(\cdot;\theta)]\|^2 & \text{if iTSONN.} \end{cases}$$

4:         (b) Update the parameters $\theta$ via gradient descent
             $\theta \leftarrow \theta - \eta\nabla_\theta \mathcal{L}(\theta)$
5:     end
6:     $\vartheta_{n+1} = \theta$
7: end

**Output:** $\hat{u}(\cdot;\theta)$

Algorithm 1 presents the unified framework of eTSONN, iTSONN and PINNs. They only difference lies in the loss function and have the same loss function value at the first step of inner iteration, but their gradients to $\theta$ are different, thus leading to different training trajectories. Note that for PINN, outer iteration is not actually needed. The outer iteration of PINN in Algorithm 1 is only for the formal uniformity of the three methods, so the result of PINN only depends on total number of iterations $N*K$. Because the TSONN does not directly minimize the residual of the PDE, its convergence is enforced by making the training follows pseudo time-stepping.

## 3 Results

In this section, TSONN is used to solve several benchmark problems where standard PINN fails. Throughout all benchmarks, we will employ the fully connected DNN architecture equipped with hyperbolic tangent activation functions (tanh). All training took on an Nvidia 4090 GPU. After training the model, we measure the L2 relative between the predicted solution and the reference solution. The relative $L_2$ errors is $\|\hat{\boldsymbol{u}} - \boldsymbol{u}_{ref}\|_2 / \|\boldsymbol{u}_{ref}\|_2$. The code and data accompanying this manuscript will be made publicly available at https://github.com/Cao-WenBo/TSONN.

### 3.1 Laplace' equation

We first consider a two-dimensional Laplace's equation for solving irrotational, incompressible flow past a cylinder with a radius $R_{wall} = 0.5$, which is the most famous and extensively studied equations in mathematical physics.

$$\frac{\partial^2 \phi}{\partial x^2} + \frac{\partial^2 \phi}{\partial y^2} = 0 \tag{10}$$

Equation (10) is the Laplace's equation and also the velocity potential equation for incompressible fluid, where $\phi$ is velocity potential function and $\boldsymbol{V} = (u,v) = (\phi_x, \phi_y)$ is the velocity vector. As shown in Figure 1(a), the flow approaches the uniform freestream conditions far away from the body. Hence, the boundary conditions on velocity in the far field are

$$\frac{\partial \phi}{\partial x} = u = V_\infty, \frac{\partial \phi}{\partial y} = v = 0 \tag{11}$$

Because the flow cannot penetrate the wall, the velocity vector must be tangent to the surface, and then the component of velocity normal to the surface must be zero. Let $\boldsymbol{n}$ be a unit vector normal to the surface as shown in Figure 1(a). The wall boundary condition can be written as

$$\nabla \phi \cdot \boldsymbol{n} = \boldsymbol{V} \cdot \boldsymbol{n} = 0. \tag{12}$$

The analytical solution of velocity is:

$$u = \frac{\partial \phi}{\partial x} = V_\infty [1 - \frac{R^2(x^2 - y^2)}{(x^2 + y^2)^2}], v = \frac{\partial \phi}{\partial y} = -V_\infty [\frac{2R^2 xy}{(x^2 + y^2)^2}] \tag{13}$$

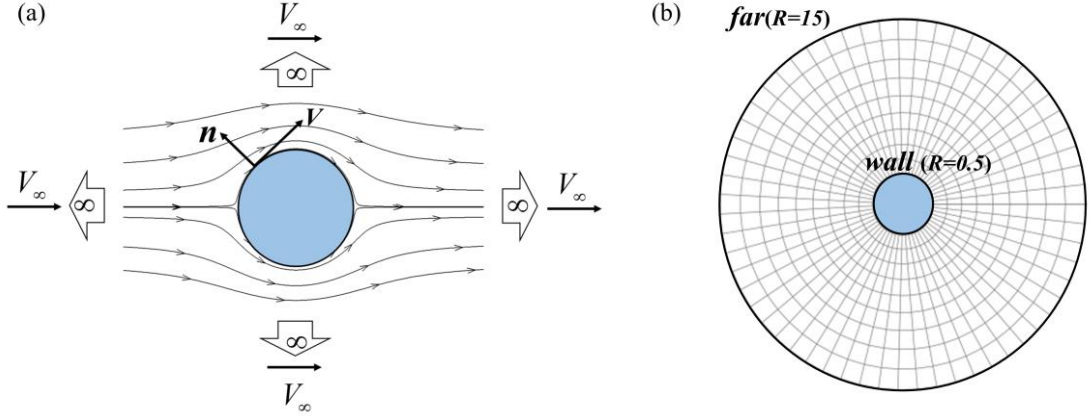

Figure 1. (a) Boundary conditions at infinity and on the wall. (b) Actual computational domain.

We solve this problem in a finite circular domain with a radius $R_{far} = 15$, and represent the velocity potential $\phi$ by a network with 5 hidden layers and 128 neurons per hidden layer. The boundary conditions on the far and wall are considered as soft constraints, and the weight $\lambda_{BC} = 0.1$. For simplicity, we create a O-mesh of size 200 × 100 in the polar computational domain $(0, 2\pi) \times (0.5, 15)$, as shown in Figure 1(b).

We train the network via full-batch gradient descent using the Adam optimizer for $5 \times 10^4$ total iterations. As shown in Figure 2 and 3, standard PINN fails to achieve stable training and obtain any meaningful results near the wall. Its loss rapidly decreases in the first few hundred training iterations, and then barely change for the rest of training, implying that the neural network gets trapped in an erroneous local minimum. In iTSONN, when the pseudo time step $\Delta \tau$ is 100, it obtains similar training history and incorrect results as PINN, indicating that PINN is a special case of iTSONN when $\Delta \tau$ is large. When $\Delta \tau \leq 10$, the loss of iTSONN can obtain stable decrease and correct result by following the trajectory of time-stepping.

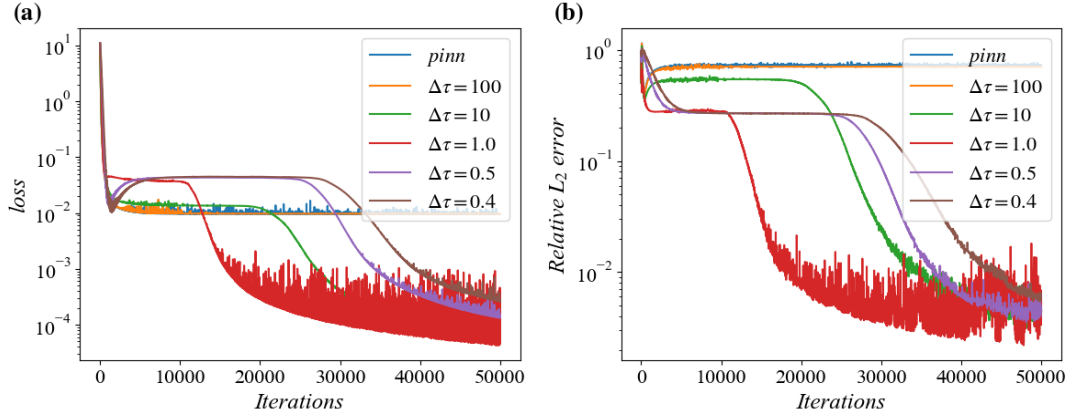

Figure 2. (a) Training losses and (b) predicted relative $L_2$ errors of PINN and iTSONN with $K=50$ for different $\Delta\tau$, where "Iterations" is the total number of iterations (i.e., $N*K$).

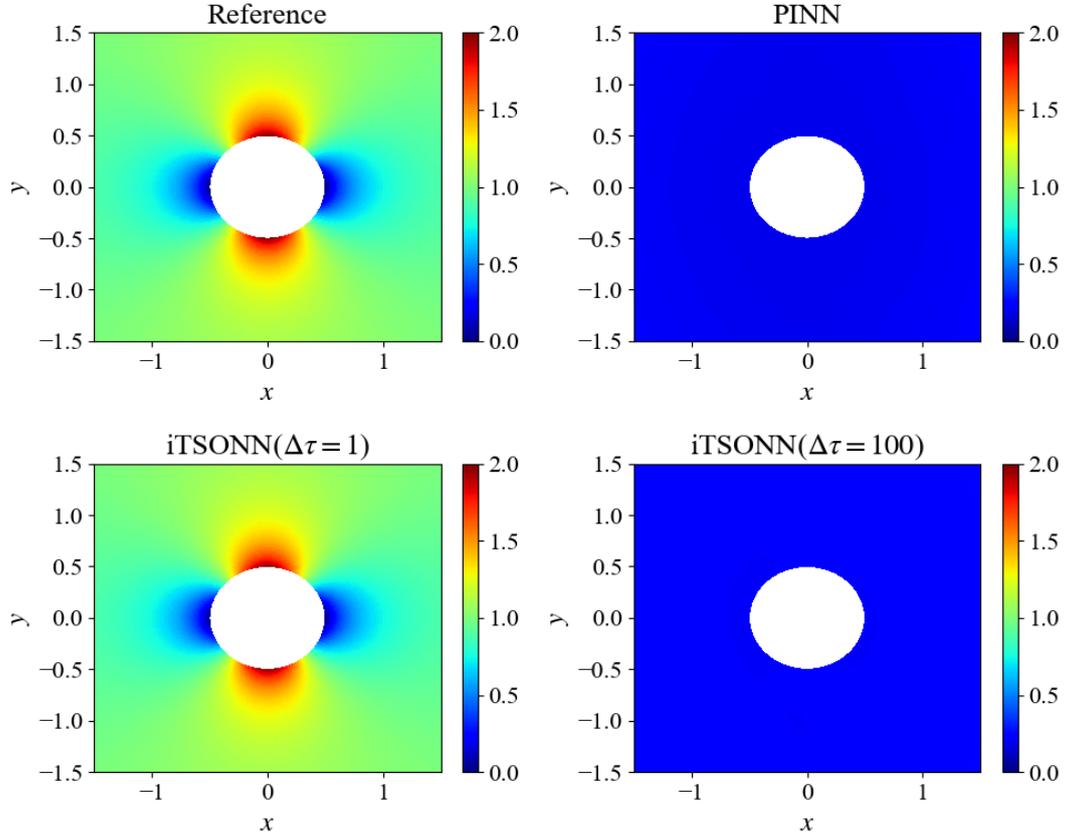

Figure 3. Contours of $u$ near the wall obtained by analytical solution, PINN, iTSONN with $\Delta\tau=1$, and iTSONN with $\Delta\tau=100$.

We further study the effect of the number of inner iterations on the training. As shown in the Figure 4, as $K$ increases, iTSONN obtains almost the same accuracy results but the convergence becomes slower. It is like traditional numerical methods that an exact solution of the implicit equation during intermediate iterations is usually time-consuming and unnecessary, but usually lead to more robust numerical schemes.

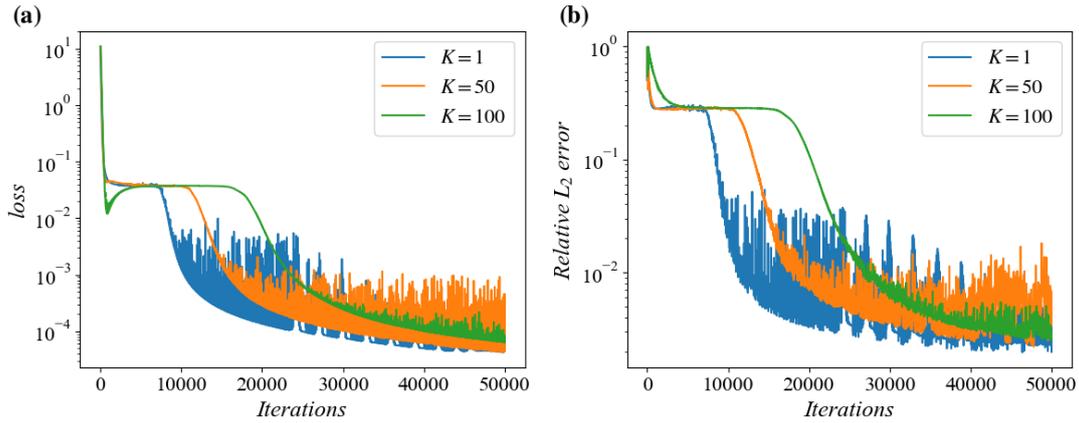

Figure 4. (a) Training losses and (b) predicted relative L$_2$ errors of iTSONN with $\Delta\tau=1$ for different *K*.

Figure 5 shows the convergence histories of eTSONN under different $\Delta\tau$. Surprisingly, eTSONN can converge even if $\Delta\tau=100$, while the maximum allowable time step in the traditional numerical method with explicit time-stepping is only about $\Delta\tau=0.0001$ under the same mesh. The explicit scheme is simple to set up, implement and easy to parallelize, but it is gradually replaced by the complex implicit scheme in CFD precisely because its time step $\Delta\tau$ is largely limited by stability constraints, thus resulting in long computer running times to reach the final steady-state. However, Figure 5 shows that explicit scheme seems not to be limited by stability constraints in NN-based optimization approach. To explain this observation, Figure 6 gives the training histories under different inner iteration number *K*. We observe the training diverges rapidly just like the traditional numerical method for lager time step when $K\geq 50$, which stems from the fact that the inner iterations enable the NN to sufficiently approximate $\bar{u}_{n+1}$ to adequately follow the trajectory of time-stepping. Therefore, the stability constraints still apply in the NN-based optimization framework, but divergence errors are naturally filtered when the inner iteration number is small, probably due to the neural network's tendency to fit low frequencies first [41] thereby introducing additional dissipation. When neural networks are used to represent the solution or the residual of PDE [42, 43], the time step beyond stability constraints is also observed in explicit time-stepping. However, further explanation is still lacking. In addition, we also observe that the convergence becomes slower as $\Delta\tau$ decreases in Figure 5, which is consistent with traditional numerical methods.

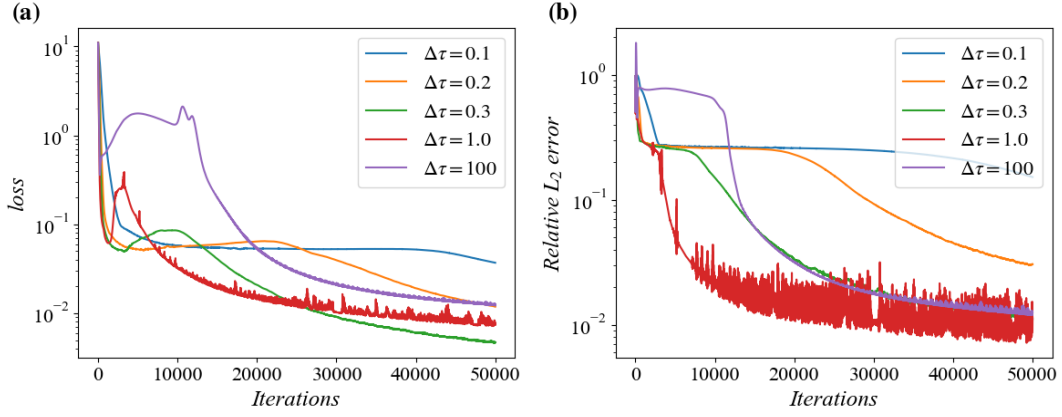

Figure 5. (a) Training losses and (b) predicted relative L$_2$ errors of eTSONN with $K$=10 for different $\Delta \tau$.

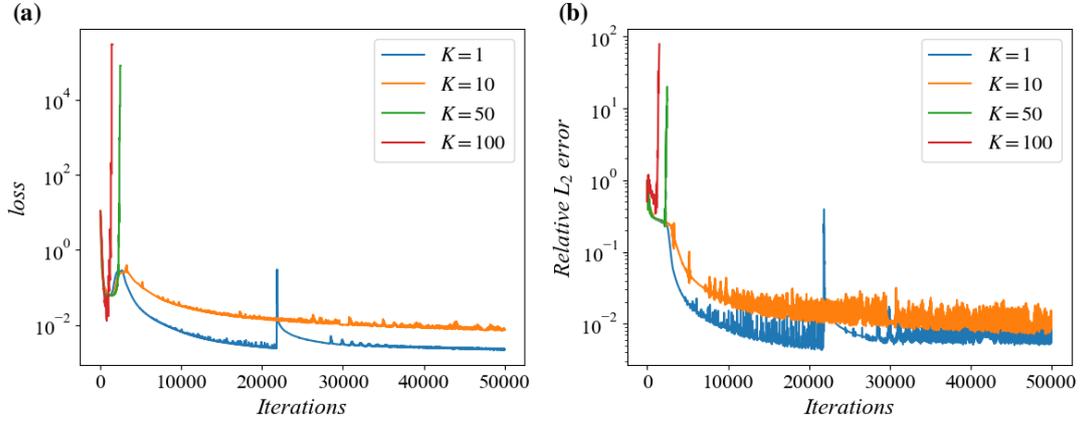

Figure 6. (a) Training losses and (b) predicted relative L2 errors of eTSONN with $\Delta \tau$=1 for different $K$.

In this case, iTSONN and eTSONN can converge even without using inner iteration (i.e., $K=1$), which is problem-dependent and not true in most cases according to our experience. We further consider the one-dimensional steady Burgers' equation (Equation (14)) to study the effect of inner iteration. It is a nonlinear partial differential equation that simulates the propagation and reflection of shock waves and takes the form $-uu_x + 0.05u_{xx} = 0$. We represent the velocity $u$ by a network with 3 hidden layers and 10 neurons per hidden layer. For simplicity, we create a uniform mesh of size 500 in the computational domain. We choose $\lambda_{BC} = 1$.

$$u\frac{\partial u}{\partial x} + 0.05\frac{\partial^2 u}{\partial x^2} = 0, x \in [-1,1]$$
$$u(-1) = 1, u(1) = -1$$
(14)

We train the network via full-batch gradient descent using the Adam optimizer.

We observed that for both eTSONN and iTSONN, when $K$ is small, they cannot obtain stable convergence and follow the trajectory of time-stepping (Figure 7) because the loss function changes too fast. In addition, eTSONN diverges due to stability constraints when $K=1000$, while iTSONN is very robust for $K \geq 10$. According to our experience, a suitable $K$ of eTSONN may be difficult to find or even non-existent in some cases. Therefore, considering that iTSONN is as easy to implement as eTSONN in the NN-based optimization approach and is more robust, we recommend using iTSONN in solving PDEs and the remaining sections of this paper only use iTSONN.

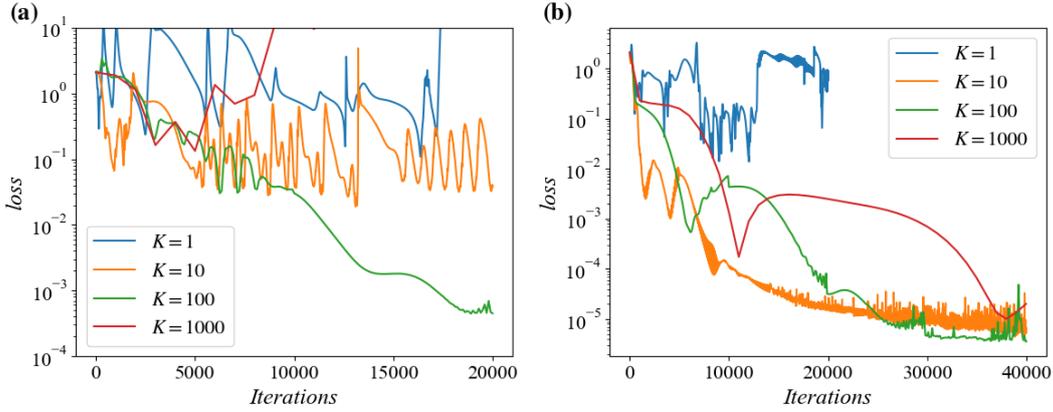

Figure 7. Training losses of (a) eTSONN and (b) iTSONN with $\Delta\tau=0.1$ for different $K$ in solving Burgers' equation.

3.2 Flow in a lid-driven cavity

We second consider a lid-driven cavity problem, which is a classical benchmark in CFD. The system is governed by the two-dimensional incompressible Navier-Stokes equations:

$$\begin{aligned} &\boldsymbol{u} \cdot \nabla \boldsymbol{u} + \nabla p - \Delta \boldsymbol{u} / \mathrm{Re} = 0 \\ &\nabla \cdot \boldsymbol{u} = 0 \\ &\boldsymbol{u} = (1,0) \quad \text{in } \Gamma_0 \\ &\boldsymbol{u} = (0,0) \quad \text{in } \Gamma_1 \end{aligned} \tag{15}$$

where $\boldsymbol{u}=(u,v)$ is velocity vector, $p$ is pressure. The computational domain $\Omega=(0,1)\times(0,1)$ is a two-dimensional square cavity, where $\Gamma_0$ is its top boundary and $\Gamma_1$ is the other three sides. Despite its simple geometry, the driven cavity flow retains a rich fluid flow physics manifested by multiple counter rotating recirculating regions on the corners of the cavity as Re increases [44]. However, it has been reported that the standard PINN fails to solve the benchmark problem when $\mathrm{Re}>100$, and some improved PINN methods only achieve the solutions for low Re ($\text{Re} < 500$) [20, 33].

We use a network with 5 hidden layers and 128 neurons per hidden layer, and train the network using LBFGS algorithm. Since the LBFGS algorithm strongly relies on the historical gradient to approximate the inverse Hessian matrix, the optimizer must to be restarted in the outer iterations for iTSONN as the loss function changes, otherwise the training diverges rapidly. Thus, iTSONN allows resampling residual points as the optimizer restarts in outer iterations to further improve robustness. The weight $\lambda_{BC}$ is set to 1. To enable the training adequately follow the trajectory of time-stepping, we choose $\Delta\tau = 0.5$ and $K = 300$. We set $N = 300$ when $\text{Re} \leq 1000$, $N = 3000$ at $\text{Re} = 2500$, and $N = 5000$ at $\text{Re} = 5000$. We enforce the PDE residuals and boundary conditions on 20,000 random residual points and 2,000 uniform boundary points, respectively, and evaluate the relative $L_2$ error on a 500*500 uniform mesh.

Figure 8 and 9 show the results of PINN and iTSONN under different Re. We observe that PINN fails to obtain correct results when $\text{Re} > 250$, while iTSONN can obtain correct results even with Re is as high as 5000. This is remarkable for NN-based optimization approach in solving PDEs, enabling them to solve more complex engineering problems.

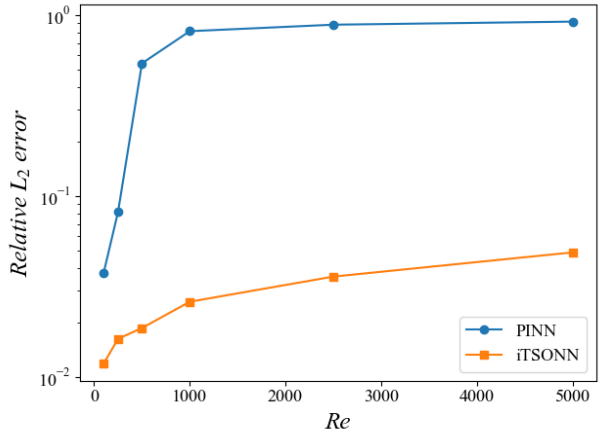

Figure 8. Predicted errors in solving lid-driven cavity flow with respect to Re.

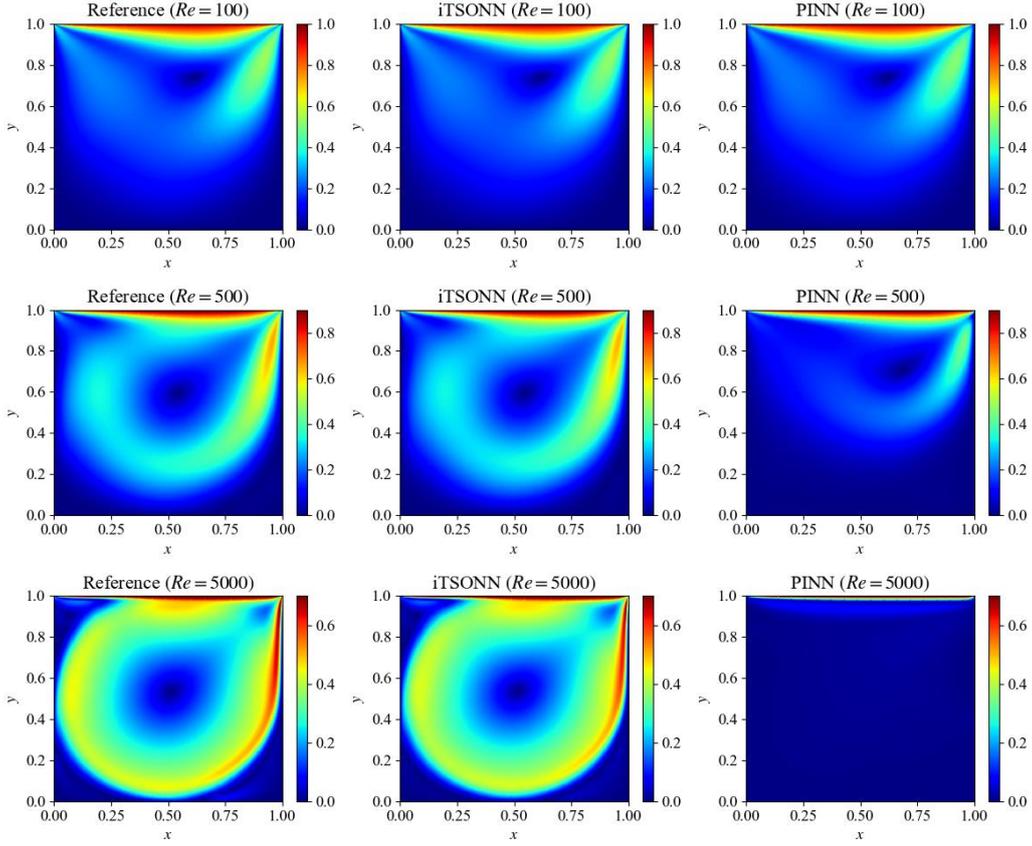

Figure 9. Contours of velocity magnitude for different Re and different methods.

Figure 10 shows the loss and error histories of PINN and iTSONN, we observe that PINN converges faster than iTSONN when Re = 100 because minimizing the PDE residual is more straightforward than the pseudo time-stepping. However, when Re > 100, the results of PINN have large errors even with a small loss function value, which verifies the ill-conditioning of the loss function [34]. On the contrary, iTSONN achieves stable convergence in both the loss and the error.

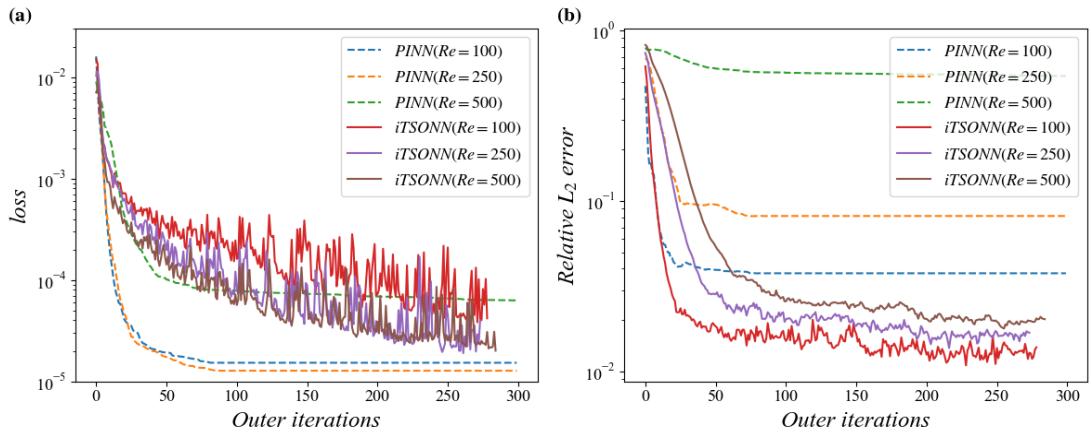

Figure 10. (a) Training losses and (b) predicted relative L2 errors for different Re.

Figure 11 shows the error histories of iTSONN against wall time. We observe that training converges slower as Re increases. It only takes 180s for the error to drop to 2.2e-2 at $Re=100$, while it takes 5220s to reach 1e-1 at $Re=5000$. Therefore, for high Reynolds number problems, the efficiency of TSONN still needs to be improved.

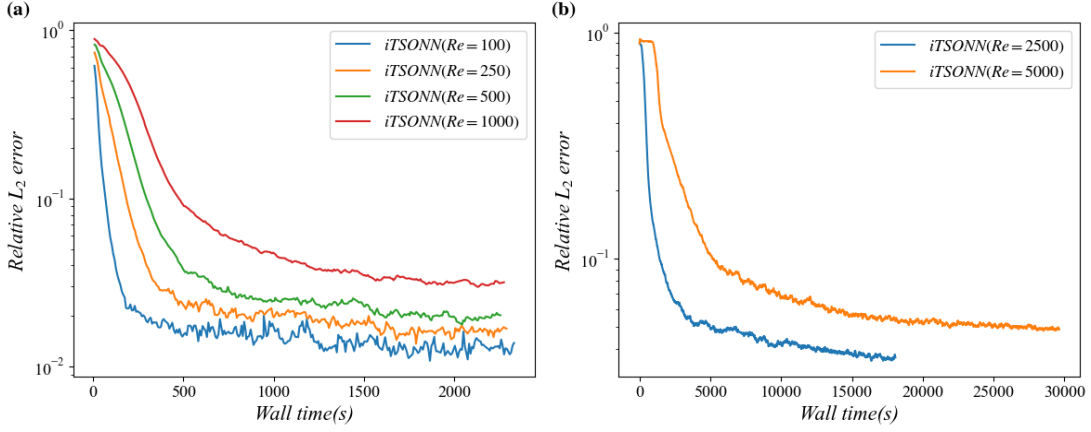

Figure 11. Predicted relative L2 errors with respect to wall time for different Re

3.3 One-dimensional Allen-Cahn equation

Next, we turn our attention to a time-dependent problem, the one-dimensional Allen-Cahn equation, which is difficult to directly solve with the standard PINNs. This example has been used in several studies to improve the performance of PINNs [45-47].

$$\begin{aligned}
&\frac{\partial u}{\partial t} - 0.0001\frac{\partial^2 u}{\partial x^2} + 5u^3 - 5u = 0, x \in [-1,1], t \in [0,1]\\
&u(x,0) = x^2 \cos(\pi x)\\
&u(t,-1) = u(t,1),\\
&\frac{\partial u}{\partial x}(t,-1) = \frac{\partial u}{\partial x}(t,1)
\end{aligned} \quad (16)$$

Following the setup discussed in these studies, we use a network with 4 hidden layers and 128 neurons per hidden layer. We choose $\lambda_{IC}=10, \lambda_{BC}=1$ and $\Delta\tau=0.3$. We enforce the PDE residuals, initial conditions and boundary conditions on 20,000 random residual points, 257 uniform initial points, and 202 uniform boundary points, respectively, and evaluate the relative $L_2$ error on a 257*101 uniform mesh. As shown in Figure 12 and 12, the PINN fails to capture the sharper transitions and obtain the correct result, while the iTSONN achieves an excellent agreement with the reference solution under the same hyper-parameter settings, yielding a relative $L_2$ error of 4.9e-03. The results show that iTSONN works well for time-dependent problems. We emphasize again that such remarkable improvements only require a simple modification on the loss function compared to PINNs.

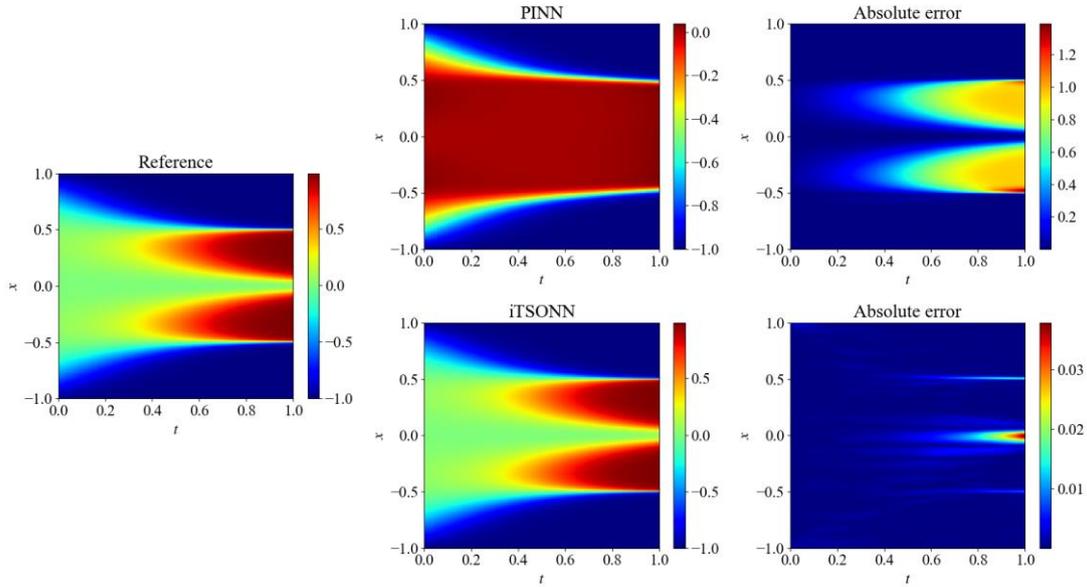

Figure 12. Reference solution and predicted solutions of PINN and iTSONN.

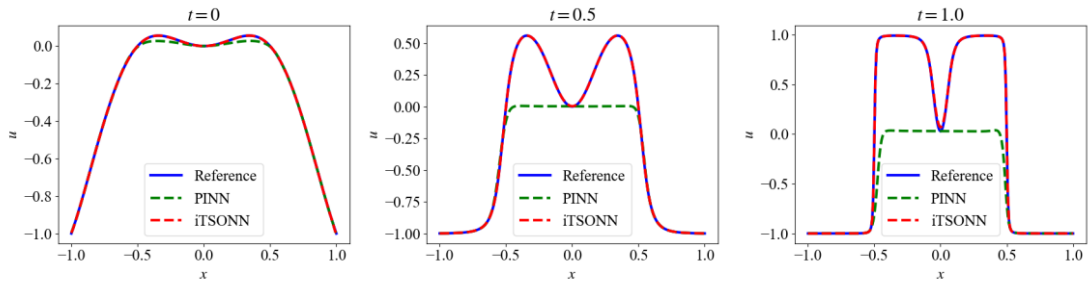

Figure 13. Comparison of the predicted and reference solutions corresponding to the three temporal snapshots at t = 0.0, 0.5, 1.0.

Figure 14 shows the relative $L_2$ error against wall time. The error decreases rapidly to about 1e-2 in the first 60 seconds, and then enters a stable and slow decline.

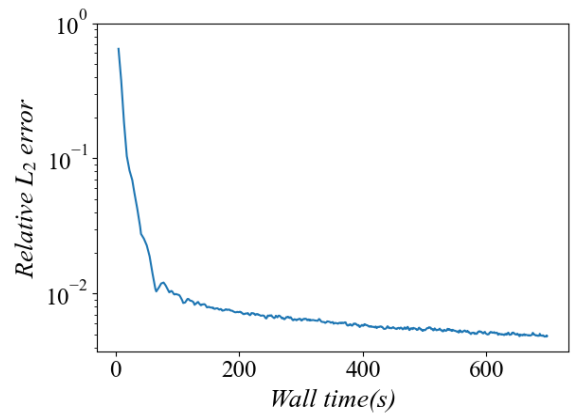

Figure 14. Predicted relative L2 error with respect to wall time.

## 4 Conclusions

In this paper, we propose the time-stepping-oriented neural network (TSONN) as a novel approach for solving partial differential equations. The TSONN integrates the time-stepping method with the deep learning, effectively enforcing the convergence of model training by following the trajectory of pseudo time-stepping. In the explicit TSONN, the loss function is only the mean square error of the network output and the label from the pseudo time-stepping, eliminating the need for PDE-related information, thus significantly reducing the ill-conditioning of the optimization problem. In the implicit TSONN, the PINN is a special case when the pseudo time step is large enough. Therefore, compared with the PINN, implicit TSONN divides the original optimization problem into a series of well-conditioned sub-problems. Our results show that TSONN robustly obtains stable training and correct results in various problems that standard PINNs fail to solve. These improvements require only a simple modification of the loss function compared to PINN. As a notable example, PINN fails to solve the lid-driven flow problem beyond Re=250, while TSONN successfully solves it even at Re=5000.

More interestingly, we highlight several novel properties and advantages of time-stepping methods within the framework of neural network-based optimization approach. Specifically, the explicit time-stepping scheme allows for significantly larger time steps, surpassing the limitations of traditional mesh-based numerical methods by several orders of magnitude. The implicit time-stepping scheme can be implemented as straightforwardly as explicit scheme without local linearization of PDEs and solution of a large system of sparse linear equations as in traditional methods.

## Data Availability

Enquiries about data availability should be directed to the authors.

## Competing interests

The authors have not disclosed any competing interests.

## Acknowledgments

We would like to acknowledge the support of the National Natural Science Foundation of China (No. 92152301). We also thank to Jiaqing Kou and Xianglin Shan for their valuable comments, which greatly contributed to improving the quality of our paper.